\documentclass{article}

\usepackage{ml98}
\usepackage{times}
\usepackage{amsmath}
\usepackage{amsthm}
\usepackage{amssymb}
\usepackage[dvips]{graphicx}
\usepackage{epic}
\usepackage{eepic}
\usepackage{ecltree}

\renewcommand{\glossary}[2]{}
\def\addcontentsline#1#2#3{}

\newcommand{\be}{\begin{equation*}}

\renewcommand{\[}{\left[}
\renewcommand{\]}{\right]}

\renewcommand{\hbar}{{\overline h}}

\newcommand{\Jt}{{\tilde{J}}}
\newcommand{\Js}{{J^*}}

\newcommand{\R}{{\mathbb R}}

\newcommand{\argmin}{{\mbox{argmin}}}

\title{KnightCap: A chess program that learns by combining 
TD($\lambda$) with game-tree search}

\author{{\bf Jonathan Baxter} \\
Department of Systems Engineering \\
Australian National University \\
Canberra 0200, Australia \\
Jon.Baxter@anu.edu.au \\
\And
{\bf Andrew Tridgell} \\
Department of Computer Science \\
Australian National University \\
Canberra 0200, Australia \\
Andrew.Tridgell@anu.edu.au \\
\And
{\bf Lex Weaver} \\
Department of Computer Science \\
Australian National University \\
Canberra 0200, Australia \\
Lex.Weaver@anu.edu.au}

\begin{document}
\maketitle

\begin{abstract}
In this paper we present TDLeaf($\lambda$), a variation on the
TD($\lambda$) algorithm that enables it to be used in conjunction with
game-tree search. We present some experiments in which our chess
program ``KnightCap'' used TDLeaf($\lambda$) to learn its evaluation
function while playing on the Free Internet Chess Server (FICS, {\tt
fics.onenet.net}). The main success we report is that KnightCap
improved from a 1650 rating to a 2150 rating in just 308 games and 3
days of play. As a reference, a rating of 1650 corresponds to about
level B human play (on a scale from E (1000) to A (1800)), while 2150
is human master level. We discuss some of the reasons for this success,
principle among them being the use of on-line, rather than self-play.
\end{abstract}

\section{Introduction}
Temporal Difference learning, first introduced by Samuel
\cite{samuel59} and later extended and formalized by Sutton
\cite{sutton88} in his TD($\lambda$) algorithm, is an elegant
technique for approximating the expected long term future cost (or
{\em cost-to-go}) of a stochastic dynamical system as a function of
the current state. The mapping from states to future cost is
implemented by a parameterized function approximator such as a neural
network. The parameters are updated online after each state
transition, or possibly in batch updates after several state
transitions. The goal of the algorithm is to improve the cost
estimates as the number of observed state transitions and associated
costs increases.

Perhaps the most remarkable success of TD($\lambda$) is Tesauro's
TD-Gammon, a neural network backgammon player that was trained from
scratch using TD($\lambda$) and simulated self-play. TD-Gammon is
competitive with the best human backgammon players \cite{Tesauro94}. In
TD-Gammon the neural network played a dual role, both as a predictor of
the expected cost-to-go of the position and as a means to select
moves. In any position the next move was chosen greedily by evaluating
all positions reachable from the current state, and then selecting the
move leading to the position with smallest expected cost. The
parameters of the neural network were updated according to the
TD($\lambda$) algorithm after each game.

Although the results with backgammon are quite striking, there is
lingering disappointment that despite several attempts, they have not
been repeated for other board games such as othello, Go and the
``drosophila of AI'' --- chess \cite{thrun95b,walker93,schraudolph94}.

Many authors have discussed the peculiarities of backgammon that make
it particularly suitable for Temporal Difference learning with
self-play \cite{tesauro92,schraudolph94,pollack96}. Principle among
these are {\em speed of play}: TD-Gammon learnt from several hundred
thousand games of self-play, {\em representation smoothness}: the
evaluation of a backgammon position is a reasonably smooth function of
the position (viewed, say, as a vector of piece counts), making it
easier to find a good neural network approximation, and {\em
stochasticity}: backgammon is a random game which forces at least a
minimal amount of exploration of search space.

As TD-Gammon in its original form only searched one-ply ahead, we feel
this list should be appended with: {\em shallow search is good enough
against humans}. There are two possible reasons for this; either one
does not gain a lot by searching deeper in backgammon (questionable
given that recent versions of TD-Gammon search to three-ply and this
significantly improves their performance), or humans are simply
incapable of searching deeply and so TD-Gammon is only competing in a
pool of shallow searchers. Although we know of no psychological
studies investigating the depth to which humans search in backgammon,
it is plausible that the combination of high branching factor and
random move generation makes it quite difficult to search more than
one or two-ply ahead. In particular, random move generation
effectively prevents selective search or ``forward pruning'' because
it enforces a lower bound on the branching factor at each move.

In contrast, finding a representation for chess, othello or Go which
allows a small neural network to order moves at one-ply with near
human performance is a far more difficult task. It seems that for
these games, reliable tactical evaluation is difficult to achieve
without deep lookahead. As deep lookahead invariably involves some
kind of minimax search, which in turn requires an exponential increase
in the number of positions evaluated as the search depth increases,
the computational cost of the evaluation function has to be low,
ruling out the use of expensive evaluation functions such as neural
networks. Consequently most chess and othello programs use linear
evaluation functions (the branching factor in Go makes minimax search
to any significant depth nearly infeasible).

In this paper we introduce TDLeaf($\lambda$), a variation on the
TD$(\lambda)$ algorithm that can be used to learn an evaluation
function for use in deep minimax search. TDLeaf($\lambda$) is
identical to TD($\lambda$) except that instead of operating on the
positions that occur during the game, it operates on the leaf nodes of
the {\em principal variation} of a minimax search from each position
(also known as the {\em principal leaves}).

To test the effectiveness of TDLeaf($\lambda$), we incorporated it
into our own chess program---{\em KnightCap}.  KnightCap has a
particularly rich board representation enabling relatively fast
computation of sophisticated positional features, although this is
achieved at some cost in speed (KnightCap is about 10 times slower
than {\em Crafty}---the best public-domain chess program---and 6,000
times slower than {\em Deep Blue}). We trained KnightCap's linear
evaluation function using TDLeaf($\lambda$) by playing it on the Free
Internet Chess Server (FICS, {\tt fics.onenet.net}) and on the
Internet Chess Club (ICC, {\tt chessclub.com}). Internet play was used
to avoid the premature convergence difficulties associated
self-play\footnote{Randomizing move choice is another way of avoiding
problems associated with self-play (this approach has been tried in Go
\cite{schraudolph94}), but the advantage of the Internet is that more
information is provided by the opponents play.}.The main success story
we report is that starting from an evaluation function in which all
coefficients were set to zero except the values of the pieces,
KnightCap went from a 1650-rated player to a 2150-rated player in just
three days and 308 games.  KnightCap is an ongoing project with new
features being added to its evaluation function all the time. We use
TDLeaf($\lambda$) and Internet play to tune the coefficients of these
features.

The remainder of this paper is organized as follows. In section
\ref{oldsec} we describe the TD($\lambda$) algorithm as it applies to
games. The TDLeaf($\lambda$) algorithm is described in section
\ref{newsec}. Experimental results for internet-play with KnightCap
are given in section \ref{expsec}. Section \ref{conclusion} contains
some discussion and concluding remarks.

\section{The TD($\lambda)$ algorithm applied to games}
\label{oldsec}
In this section we describe the TD($\lambda$) algorithm as it applies
to playing board games. We discuss the algorithm from the point of
view of an {\em agent} playing the game.

Let $S$ denote the set of all possible board positions in the game.
Play proceeds in a series of moves at discrete time steps
$t=1,2,\dots$. At time $t$ the agent finds itself in some position
$x_t\in S$, and has available a set of moves, or {\em
actions} $A_{x_t}$ (the legal moves in position $x_t$). The agent
chooses an action $a\in A_{x_t}$ and makes a transition to state
$x_{t+1}$ with probability $p(x_t,x_{t+1},a)$.
Here $x_{t+1}$ is the position of the board after the agent's move and the
opponent's response.  When the game is over, the agent receives a
scalar reward, typically ``1'' for a win, ``0'' for a draw and ``-1''
for a loss.

For ease of notation we will assume all games have a fixed length of
$N$ (this is not essential). Let $r(x_N)$ denote the reward received
at the end of the game. If we assume that the agent chooses its
actions according to some function $a(x)$ of the current state $x$ (so
that $a(x) \in A_{x}$), the expected reward from each state $x\in S$ is given
by 
\begin{equation}
\label{expreward}
\Js(x) := E_{x_N|x} r(x_N),
\end{equation}
where the expectation is with respect to the transition probabilities
$p(x_t,x_{t+1}, a(x_t))$ and possibly also with respect to the actions
$a(x_t)$ if the agent chooses its actions stochastically. 

For very large state spaces $S$ it is not possible store the value of
$\Js(x)$ for every $x\in S$, so instead we might try to approximate
$J^*$ using a parameterized function class $\Jt\colon S\times \R^k \to
\R$, for example linear function, splines, neural networks,
etc. $\Jt(\cdot,w)$ is assumed to be a differentiable function of its
parameters $w=(w_1,\dots,w_k)$. The aim is to find a parameter vector
$w\in \R^k$ that minimizes some measure of error between the
approximation $\Jt(\cdot,w)$ and $\Js(\cdot)$. The TD($\lambda$)
algorithm, which we describe now, is designed to do exactly that.

Suppose $x_1,\dots,x_{N-1}, x_N$ is a sequence of states in one game.
For a given parameter vector $w$, define the {\em temporal difference}
associated with the transition $x_t\rightarrow x_{t+1}$ by
\begin{equation}
\label{tempdiff}
d_t := \Jt(x_{t+1},w) - \Jt(x_t,w).
\end{equation}
Note that $d_t$ measures the difference between the reward predicted
by $\Jt(\cdot,w)$ at time $t+1$, and the reward predicted by
$\Jt(\cdot,w)$ at time $t$. The true evaluation function $\Js$ has the
property
$$
E_{x_{t+1}|x_t} \[\Js(x_{t+1}) - \Js(x_t)\] = 0,
$$
so if $\Jt(\cdot,w)$ is a good approximation to $\Js$, 
$E_{x_{t+1}|x_t} d_t$ should be close to zero. 
For ease of notation we will assume that $\Jt(x_N,w) = r(x_N)$ always,
so that the final temporal difference satisfies
$$
d_{N-1} = \Jt(x_{N},w) - \Jt(x_{N-1},w) =  r(x_N) -  \Jt(x_{N-1},w).
$$
That is, $d_{N-1}$ is the difference between the true outcome of the
game and the prediction at the penultimate move.  

At the end of the game, the TD($\lambda$) algorithm updates the
parameter vector $w$ according to the formula
\begin{equation}
\label{tdeq}
w := w + \alpha\sum_{t=1}^{N-1} \nabla\Jt(x_t,w)\[\sum_{j=t}^{N-1} 
\lambda^{j-t}d_t\]
\end{equation}
where $\nabla\Jt(\cdot,w)$ is the vector of partial derivatives of
$\Jt$ with respect to its parameters. The positive parameter $\alpha$
controls the learning rate and would typically be ``annealed'' towards
zero during the course of a long series of games. The parameter
$\lambda\in [0,1]$ controls the extent to which temporal differences
propagate backwards in time. To see this, compare equation
\eqref{tdeq} for $\lambda=0$:
\begin{align}
\label{td0}
w := &w + \alpha\sum_{t=1}^{N-1} \nabla\Jt(x_t,w)d_t \notag\\
   = &w + \alpha\sum_{t=1}^{N-1} \nabla\Jt(x_t,w)
\[\Jt(x_{t+1},w) - \Jt(x_t,w)\]
\end{align}
and $\lambda=1$:
\begin{equation}
\label{td1}
w := w + \alpha\sum_{t=1}^{N-1} \nabla\Jt(x_t,w)
\[r(x_N) - \Jt(x_t,w)\].
\end{equation}
Consider each term contributing to the sums in equations \eqref{td0}
and \eqref{td1}.  For $\lambda=0$ the parameter vector is being
adjusted in such a way as to move $\Jt(x_t,w)$---the predicted reward
at time $t$---closer to $\Jt(x_{t+1},w)$---the predicted reward at
time $t+1$.  In contrast, TD(1) adjusts the parameter vector in such
away as to move the predicted reward at time step $t$ closer to the
final reward at time step $N$. Values of $\lambda$ between zero and one
interpolate between these two behaviors. Note that \eqref{td1} is
equivalent to gradient descent on the error function $E(w) :=
\sum_{t=1}^{N-1}\[r(x_N) - \Jt(x_t,w)\]^2$. 

Successive parameter updates according to the TD($\lambda$) algorithm
should, over time, lead to improved predictions of the expected reward
$\Jt(\cdot, w)$. Provided the actions $a(x_t)$ are independent of the
parameter vector $w$, it can be shown that for {\em linear}
$\Jt(\cdot,w)$, the TD($\lambda$) algorithm converges to a
near-optimal parameter vector \cite{tsitsiklis97}. Unfortunately,
there is no such guarantee if $\Jt(\cdot,w)$ is non-linear
\cite{tsitsiklis97}, or if $a(x_t)$ depends on $w$
\cite{Bertsekas96}.

\section{Minimax Search and TD($\lambda$)}
\label{newsec}
\begin{figure}
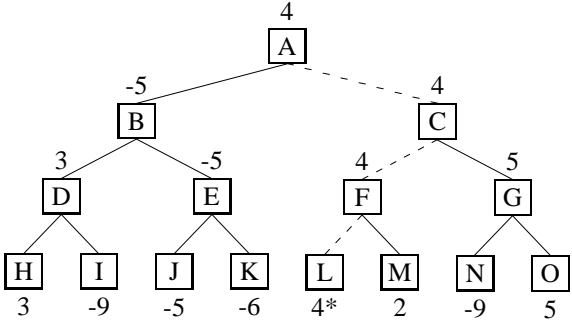

\begin{center}
\begin{bundle}{\shortstack{4 \\ \framebox[5mm]{A}}}
\offinterlineskip
\drawwith{\drawwith{\dashline{3}}\drawline}     
\chunk[-5]{
        \begin{bundle}{\framebox[5mm]{B}}
        \drawwith{\drawline}
        \chunk[3]{
                \begin{bundle}{\framebox[5mm]{D}}
                \drawwith{\drawline}
                \chunk{\shortstack{\framebox[5mm]{H} \\ 3}}
                \chunk{\shortstack{\framebox[5mm]{I} \\ -9}}
                \end{bundle}}
        \chunk[-5]{
                \begin{bundle}{\framebox[5mm]{E}}
                \chunk{\shortstack{\framebox[5mm]{J} \\ -5}}
                \chunk{\shortstack{\framebox[5mm]{K} \\ -6}}
                \end{bundle}}
        \end{bundle}}
\chunk[4]{
        \begin{bundle}{\framebox[5mm]{C}}
        \drawwith{\drawwith{\drawline}\dashline{3}}
        \chunk[4]{
                \begin{bundle}{\framebox[5mm]{F}}
                \chunk{\shortstack{\framebox[5mm]{L} \\ 4*}}
                \chunk{\shortstack{\framebox[5mm]{M} \\ 2}}
                \end{bundle}}
        \chunk[5]{
                \begin{bundle}{\framebox[5mm]{G}}
                \drawwith{\drawline}
                \chunk{\shortstack{\framebox[5mm]{N} \\ -9}}
                \chunk{\shortstack{\framebox[5mm]{O} \\ 5}}
                \end{bundle}}
        \end{bundle}}
\end{bundle}
\end{center}
\caption{Full breadth, 3-ply search tree illustrating the minimax rule
for propagating values. Each of the leaf nodes (H--O) is given a score
by the evaluation function, $\Jt(\cdot, w)$. These scores are then
propagated back up the tree by assigning to each opponent's internal
node the minimum of its children's values, and to each of our internal
nodes the maximum of its children's values. The principle variation is
then the sequence of best moves for either side starting from the root
node, and this is illustrated by a dashed line in the figure. Note
that the score at the root node A is the evaluation of the leaf node
(L) of the principal variation. As there are no ties between any
siblings, the derivative of A's score with respect to the parameters $w$
is just $\nabla\Jt(L, w)$.
\label{minimaxfig}}
\end{figure}

For argument's sake, assume any action $a$ taken in state $x$ leads to
predetermined state which we will denote by $x'_a$. Once an
approximation $\Jt(\cdot,w)$ to $\Js$ has been found, we can use it to
choose actions in state $x$ by picking the action $a\in A_x$ whose
successor state $x'_a$ minimizes the opponent's expected
reward\footnote{If successor states are only determined stochastically
by the choice of $a$, we would choose the action minimizing the
expected reward over the choice of successor states.}:
\begin{equation}
\label{actioneq}
a^*(x) := \argmin_{a\in A_x} \Jt(x'_a, w).
\end{equation}
This was the strategy used in TD-Gammon. Unfortunately, for games like
othello and chess it is very difficult to accurately evaluate a
position by looking only one move or {\em ply} ahead. Most programs
for these games employ some form of {\em minimax} search. In minimax
search, one builds a tree from position $x$ by examining all possible
moves for the computer in that position, then all possible moves for
the opponent, and then all possible moves for the computer and so on
to some predetermined depth $d$. The leaf nodes of the tree are then
evaluated using a heuristic evaluation function (such as
$\Jt(\cdot,w)$), and the resulting scores are propagated back up the
tree by choosing at each stage the move which leads to the best
position for the player on the move.  See figure \ref{minimaxfig} for
an example game tree and its minimax evaluation. With reference to the
figure, note that the evaluation assigned to the root node is the
evaluation of the leaf node of the {\em principal variation}; the
sequence of moves taken from the root to the leaf if each side chooses
the best available move.

\setlength{\GapWidth}{5mm}

\setlength{\GapWidth}{5mm}
\begin{figure}
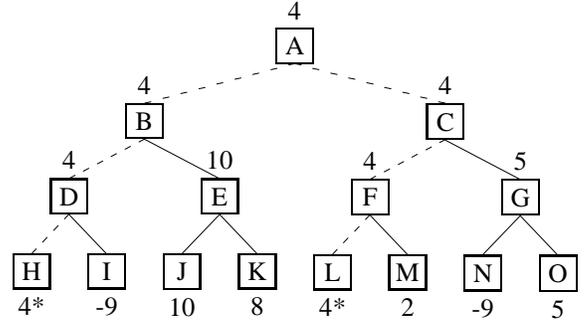

\begin{center}
\begin{bundle}{\shortstack{4 \\ \framebox[5mm]{A}}}
\offinterlineskip
\drawwith{\dashline{3}}
\chunk[4]{
        \begin{bundle}{\framebox[5mm]{B}}
        \drawwith{\drawwith{\drawline}\dashline{3}}
        \chunk[4]{
                \begin{bundle}{\framebox[5mm]{D}}
        \drawwith{\drawwith{\drawline}\dashline{3}}
                \chunk{\shortstack{\framebox[5mm]{H} \\ 4*}}
                \chunk{\shortstack{\framebox[5mm]{I} \\ -9}}
                \end{bundle}}
        \chunk[10]{
                \begin{bundle}{\framebox[5mm]{E}}
                \drawwith{\drawline}
                \chunk{\shortstack{\framebox[5mm]{J} \\ 10}}
                \chunk{\shortstack{\framebox[5mm]{K} \\ 8}}
                \end{bundle}}
        \end{bundle}}
\chunk[4]{
        \begin{bundle}{\framebox[5mm]{C}}
        \drawwith{\drawwith{\drawline}\dashline{3}}
        \chunk[4]{
                \begin{bundle}{\framebox[5mm]{F}}
                \chunk{\shortstack{\framebox[5mm]{L} \\ 4*}}
                \chunk{\shortstack{\framebox[5mm]{M} \\ 2}}
                \end{bundle}}
        \chunk[5]{
                \begin{bundle}{\framebox[5mm]{G}}
                \drawwith{\drawline}
                \chunk{\shortstack{\framebox[5mm]{N} \\ -9}}
                \chunk{\shortstack{\framebox[5mm]{O} \\ 5}}
                \end{bundle}}
        \end{bundle}}
\end{bundle}
\end{center}
\caption{A search tree with a non-unique principal variation (PV).  In
this case the derivative of the root node A with respect to the
parameters of the leaf-node evaluation function is multi-valued,
either $\nabla \Jt(H,w)$ or $\nabla \Jt(L,w)$. Except for
transpositions (in which case H and L are identical and the derivative
is single-valued anyway), such ``collisions'' are
likely to be extremely rare, so in 
TDLeaf($\lambda$) we ignore them by choosing a leaf node arbitrarily 
from the available candidates.
\label{pv}}
\end{figure}

In practice many engineering tricks are used to improve the
performance of the minimax algorithm, $\alpha-\beta$ search being the
most famous.

Let $\Jt_d(x,w)$ denote the evaluation obtained for state $x$ by
applying $\Jt(\cdot,w)$ to the leaf nodes of a depth $d$ minimax
search from $x$. Our aim is to find a parameter vector $w$ such that
$\Jt_d(\cdot,w)$ is a good approximation to the expected reward
$\Js$. One way to achieve this is to apply the TD($\lambda$)
algorithm to $\Jt_d(x,w)$. That is, for each sequence of positions
$x_1,\dots,x_N$ in a game we define the temporal differences
\begin{equation}
\label{tempdiffd}
d_t := \Jt_d(x_{t+1},w) - \Jt_d(x_t,w)
\end{equation}
as per equation \eqref{tempdiff}, and then the TD($\lambda$) algorithm
\eqref{tdeq} for updating the parameter vector $w$ becomes
\begin{equation}
\label{tdeqd}
w := w + \alpha\sum_{t=1}^{N-1} \nabla\Jt_d(x_t,w)\[\sum_{j=t}^{N-1} 
\lambda^{j-t}d_t\].
\end{equation}
One problem with equation \eqref{tdeqd} is that for $d>1$,
$\Jt_d(x,w)$ is not necessarily a differentiable function of $w$ for
all values of $w$, even if $\Jt(\cdot,w)$ is everywhere
differentiable. This is because for some values of $w$ there will be
``ties'' in the minimax search, i.e.  there will be more than one best
move available in some of the positions along the principal variation,
which means that the principal variation will not be unique (see
figure \ref{pv}). Thus, the evaluation assigned to the root node,
$\Jt_d(x,w)$, will be the evaluation of any one of a number of leaf
nodes.

Fortunately, under some mild technical assumptions on the behavior of
$\Jt(x,w)$, it can be shown that for each state
$x$, the set of $w\in\R^k$ for which $\Jt_d(x,w)$ is not
differentiable has Lebesgue measure zero. Thus for all states $x$ and
for ``almost all'' $w\in \R^k$, $\Jt_d(x,w)$ is a differentiable
function of $w$.  Note that $\Jt_d(x,w)$ is also a continuous function
of $w$ whenever $\Jt(x,w)$ is a continuous function of $w$. This
implies that even for the ``bad'' pairs $(x,w)$, $\nabla\Jt_d(x,w)$ is
only undefined because it is multi-valued. Thus we can still
arbitrarily choose a particular value for $\nabla\Jt_d(x,w)$ if $w$
happens to land on one of the bad points.

Based on these observations we modified the TD($\lambda$) algorithm to
take account of minimax search in an almost trivial way: instead of
working with the root positions $x_1,\dots,x_N$, the TD($\lambda$)
algorithm is applied to the leaf positions found by minimax search
from the root positions. We call this algorithm
TDLeaf($\lambda$). Full details are given in figure \ref{tdleaffig}.

\begin{figure*}[t]
\makebox[\textwidth]{\hss%
\fbox{\parbox{\textwidth}{Let $\Jt(\cdot,w)$ be a class of evaluation functions parameterized by
$w\in\R^k$. Let $x_1,\dots,x_N$ be $N$ positions that occurred during
the course of a game, with $r(x_N)$ the outcome of the game. For
notational convenience set $\Jt(x_N,w) := r(x_N)$.
\begin{enumerate}
\item For each state $x_i$, compute $\Jt_d(x_i,w)$ by 
performing minimax search to depth $d$ from $x_i$
and using $\Jt(\cdot, w)$ to score the leaf nodes. Note that $d$ may
vary from position to position. 
\item Let $x^l_i$ denote the leaf node of the principle variation 
starting at $x_i$. If there is more than one principal variation,
choose a leaf node from the available candidates at random. Note that 
\begin{equation}
\label{leafeq}
\Jt_d(x_i,w) = \Jt(x^l_i,w).
\end{equation}
\item For $t=1,\dots,N-1$, compute the temporal differences:
\begin{equation}
\label{leaftempdiff}
d_t := \Jt(x_{t+1}^l,w) - \Jt(x^l_t,w).
\end{equation}
\item Update $w$ according to the TDLeaf($\lambda$) formula:
\begin{equation}
\label{leaftdeq}
w := w + \alpha\sum_{t=1}^{N-1} \nabla\Jt(x^l_t,w)\[\sum_{j=t}^{N-1} 
\lambda^{j-t}d_t\].
\end{equation}
\end{enumerate}}}
\hss}
\caption{The TDLeaf($\lambda$) algorithm\label{tdleaffig}} 
\end{figure*} 

\section{TDLeaf($\lambda$) and Chess}
\label{expsec}
In this section we describe the outcome of several experiments in
which the TDLeaf($\lambda$) algorithm was used to train the weights of
a linear evaluation function in our chess program ``KnightCap''.
KnightCap is a reasonably sophisticated computer chess program for
Unix systems. It has all the standard algorithmic features that modern
chess programs tend to have as well as a number of features that are
much less common. For more details on KnightCap, including the source
code, see {\tt wwwsyseng.anu.edu.au/lsg}.

\subsection{Experiments with KnightCap}
In our main experiment we took KnightCap's evaluation function and set
all but the material parameters to zero. The material parameters were
initialized to the standard ``computer'' values: 1 for a pawn, 4 for a
knight, 4 for a bishop, 6 for a rook and 12 for a queen.  With these
parameter settings KnightCap (under the pseudonym ``WimpKnight'') was
started on the Free Internet Chess server (FICS, {\tt
fics.onenet.net}) against both human and computer opponents.  We
played KnightCap for 25 games without modifying its evaluation
function so as to get a reasonable idea of its rating. After 25 games
it had a blitz (fast time control) rating of $1650\pm 50$\footnote{the
standard deviation for all ratings reported in this section is about
50}, which put it at about B-grade human performance (on a scale from
E (1000) to A (1800)), although of course the kind of game KnightCap
plays with just material parameters set is very different to human
play of the same level (KnightCap makes no short-term tactical errors
but is positionally completely ignorant). We then turned on the
TDLeaf($\lambda$) learning algorithm, with $\lambda=0.7$ and the
learning rate $\alpha=1.0$. The value of $\lambda$ was chosen
heuristically, based on the typical delay in moves before an error
takes effect, while $\alpha$ was set high enough to ensure rapid
modification of the parameters. A couple of minor modifications to the
algorithm were made:
\begin{itemize}
\item The raw (linear) leaf node evaluations $\Jt(x_i^l,w)$ were
converted to a score between $-1$ and $1$ by computing
$$v_i^l := \tanh\[\beta\Jt(x_i^l,w)\].$$ This ensured small
fluctuations in the relative values of leaf nodes did not produce
large temporal differences (the values $v_i^l$ were used in place of
$\Jt(x_i^l,w)$ in the TDLeaf($\lambda$) calculations). The outcome of
the game $r(x_N)$ was set to 1 for a win, $-1$ for a loss and $0$ for
a draw. $\beta$ was set to ensure that a value of
$\tanh\[\beta\Jt(x_i^l,w)\] = 0.25$ was equivalent to a material
superiority of 1 pawn (initially).
\item The temporal differences, $d_t = v_{t+1}^l - v_t^l$, were 
modified in the following way. Negative values of $d_t$ were left
unchanged as any decrease in the evaluation from one position to the
next can be viewed as mistake. However, positive values of $d_t$ can
occur simply because the opponent has made a blunder. To avoid
KnightCap trying to learn to predict its opponent's blunders, we set all
positive temporal differences to zero unless KnightCap predicted the 
opponent's move\footnote{In a later experiment we only set positive
temporal differences to zero if KnightCap did not predict the
opponent's move {\em and} the opponent was rated less than
KnightCap. After all, predicting a stronger opponent's blunders is a useful 
skill, although whether this made any difference is not clear.}.

\item The value of a pawn was kept fixed at its initial value so as to
allow easy interpretation of weight values as multiples of the pawn
value (we actually experimented with not fixing the pawn value and
found it made little difference: after 1764 games with an adjustable 
pawn its value had fallen by less than  7 percent). 
\end{itemize}

Within 300 games KnightCap's rating had risen to 2150, an increase of
500 points in three days, and to a level comparable with human
masters.  At this point KnightCap's performance began to plateau,
primarily because it does not have an opening book and so will
repeatedly play into weak lines. We have since implemented an opening
book learning algorithm and with this KnightCap now plays at a rating
of 2400--2500 (peak 2575) on the other major internet chess server:
ICC, {\tt chessclub.com}\footnote{There appears to be a systematic
difference of around 200--250 points between the two servers, so a
peak rating of 2575 on ICC roughly corresponds to a peak of 2350 on
FICS. We transferred KnightCap to ICC because there are more strong
players playing there.} It often beats International Masters at
blitz. Also, because KnightCap automatically learns its parameters we
have been able to add a large number of new features to its evaluation
function: KnightCap currently operates with 5872 features (1468
features in four stages: opening, middle, ending and
mating\footnote{In reality there are not 1468 independent ``concepts''
per stage in KnightCap's evaluation function as many of the features
come in groups of 64, one for each square on the board (like the value
of placing a rook on a particular square, for example)}). With this
extra evaluation power KnightCap easily beats versions of Crafty
restricted to search only as deep as itself. However, a big caveat to 
all this optimistic assessment is that KnightCap routinely gets
crushed by faster programs searching more deeply. It is quite unlikely
this can be easily fixed simply by modifying the evaluation function,
since for this to work one has to be able to predict tactics
statically, something that seems very difficult to do. If one could
find an effective algorithm for ``learning to search selectively''
there would be potential for far greater improvement.

\begin{figure}
\begin{center}
\leavevmode
\includegraphics[width=3.3in]{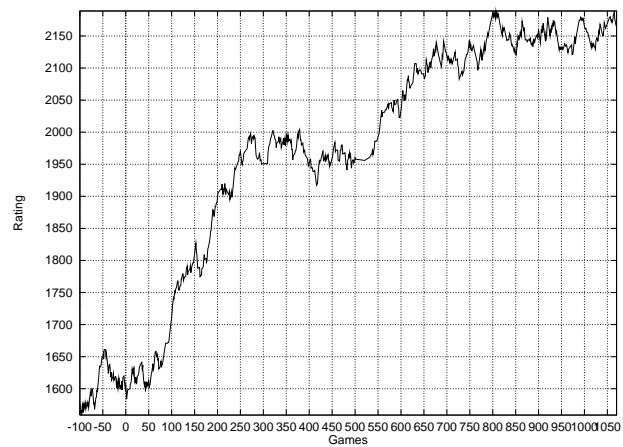}
\caption{KnightCap's rating as a function of games played (second
experiment). Learning was turned on at game 0. \label{rating}}
\end{center}
\end{figure}

Note that we have twice repeated the learning experiment and found a
similar rate of improvement and final performance level. The rating as
a function of the number of a games from one of these repeat runs is
shown in figure \ref{rating} (we did not record this information in
the first experiment). Note that in this case KnightCap took mearly
twice as long to reach the 2150 mark, but this was partly because it
was operating with limited memory (8Mb) until game 500 at which point
the memory was increased to 40Mb (KnightCap's search
algorithm---MTD(f) \cite{plaat96}---is a memory intensive variant of
$\alpha$--$\beta$ and when learning KnightCap must store the whole
position in the hash table so small memory really hurts the
performance). Another reason may also have been that for a portion of
the run we were performing paramater updates after every four games
rather than every game.

\begin{figure}
\begin{center}
\leavevmode
\includegraphics[width=3.3in]{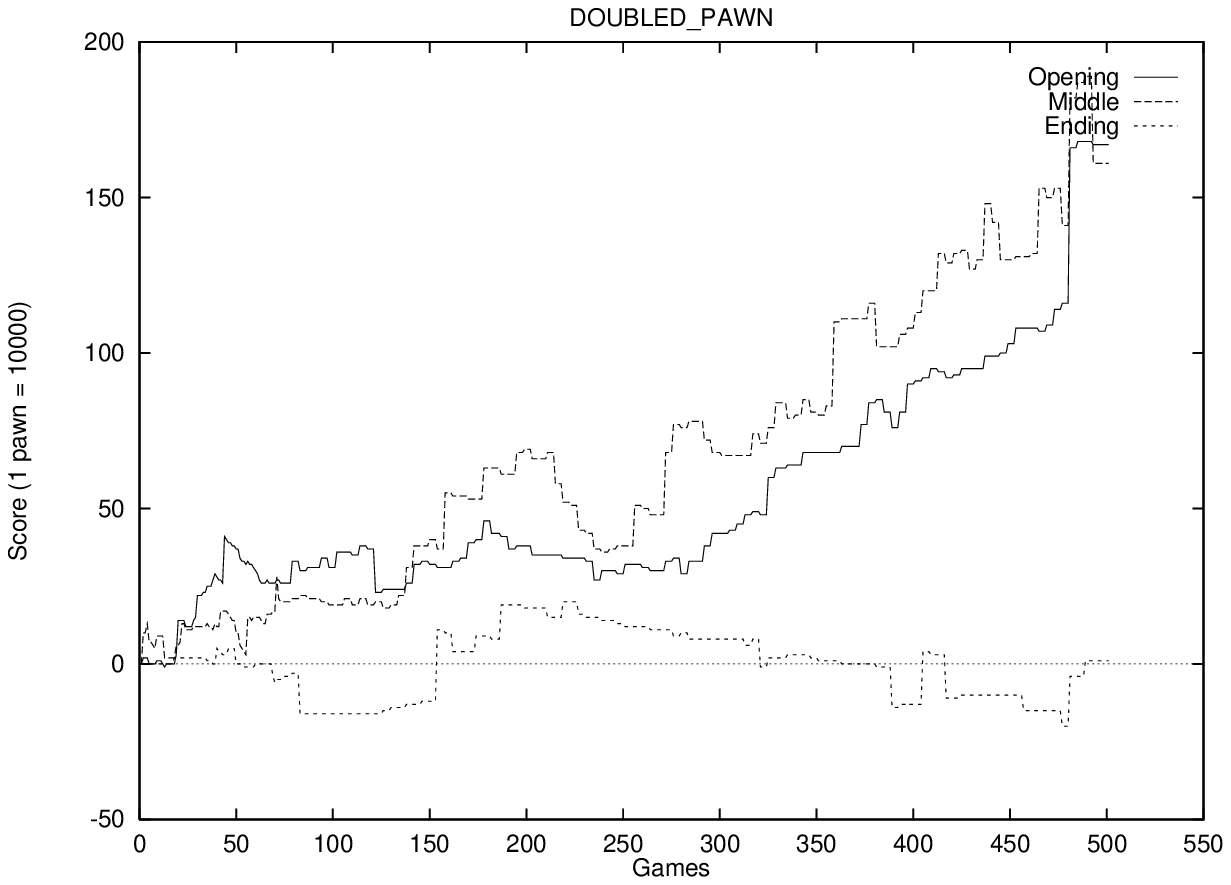} \\
\includegraphics[width=3.3in]{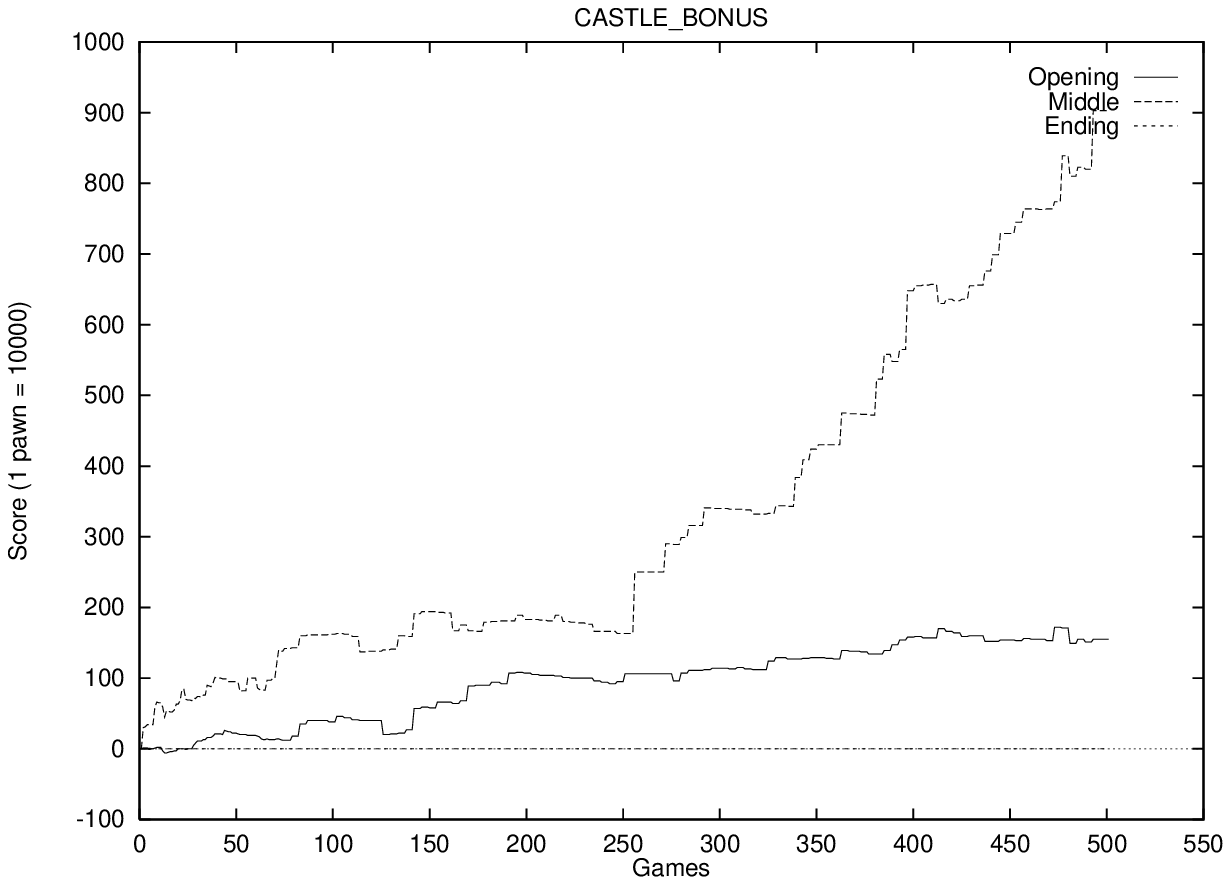}
\caption{Evolution of two paramaters (bonus for castling and penalty
for a doubled pawn) as a function of the number of games played. 
Note that each
parameter appears three times: once for each of the three stages in
the evaluation function.\label{param}}
\end{center}
\end{figure}

Plots of various parameters as a function of the number of games
played are shown in Figure \ref{param} (these plots are from the same
experiment in figure \ref{rating}). Each plot contains three graphs
corresponding to the three different stages of the evaluation
function: opening, middle and ending\footnote{KnightCap actually has a
fourth and final stage ``mating'' which kicks in when all the pieces
are off, but this stage only uses a few of the coefficients
(opponent's king mobiliity and proximity of our king to the opponent's
king).}. 

Finally, we compared the performance of KnightCap with its learnt
weight to KnightCap's performance with a set of hand-coded weights,
again by playing the two versions on ICC. The hand-coded weights were
close in performance to the learnt weights (perhaps 50-100 rating
points worse). We also tested the result of allowing KnightCap to
learn starting from the hand-coded weights, and in this case it seems
that KnightCap performs better than when starting from just material
values (peak performance was 2632 compared to 2575, but these figures
are very noisy). We are conducting more tests to verify these
results. However, it should not be too surprising that learning from a
good quality set of hand-crafted parameters is better than just
learning from material parameters. In particular, some of the
handcrafted parameters have very high values (the value of an
``unstoppable pawn'', for example) which can take a very long time to
learn under normal playing conditions, particularly if they are rarely
active in the principal leaves. It is not yet clear whether given a
sufficient number of games this dependence on the initial conditions
can be made to vanish.

\subsection{Discussion}
There appear to be a number of reasons for the remarkable rate at
which KnightCap improved. 
\begin{enumerate}
\item 
\label{p1}
As all the non-material weights were initially zero,
even small changes in these weights could cause very large changes in
the relative ordering of materially equal positions. Hence
even after a few games KnightCap was playing a substantially better
game of chess.
\item 
\label{p2}
It seems to be important that KnightCap started out life with
intelligent material parameters. This put it close in parameter space
to many far superior parameter settings. 
\item 
\label{p3}
Most players on FICS prefer to play opponents of similar strength, and
so KnightCap's opponents improved as it did. This may have had the
effect of {\em guiding} KnightCap along a path in weight space that led to a
strong set of weights. 
\item 
\label{p5} 
KnightCap was learning on-line, not by self-play. The advantage of
on-line play is that there is a great deal of information provided by
the opponent's moves. In particular, against a stronger opponent
KnightCap was being shown positions that 1) could be forced (against
KnightCap's weak play) and 2) were mis-evaluated by its evaluation
function. Of course, in self-play KnightCap can also discover
positions which are misevaluated, but it will not find the kinds of
positions that are relevant to strong play against other opponents.
In this setting, one can view the information provided by the
opponent's moves as partially solving the ``exploration'' part of the
{\em exploration/exploitation} tradeoff.
\end{enumerate}

To further investigate the importance of some of these reasons, we
conducted several more experiments. \\

\noindent {\bf Good initial conditions.} \\ A second
experiment was run in which KnightCap's coefficients were all
initialised to the value of a pawn. The value of a pawn needs to be
positive in KnightCap because it is used in many other places in the
code: for example we deem the MTD search to have converged if $\alpha
< \beta + 0.07*${\tt PAWN}. Thus, to set all parameters equal to the
same value, that value had to be a pawn.  

Playing with the initial weight settings KnightCap had a blitz rating
of around 1250.  After more than 1000 games on FICS KnightCap's rating
has improved to about 1550, a 300 point gain. This is a much slower
improvement than the original experiment. We do not know
whether the coefficients would have eventually converged to good
values, but it is clear from this experiment that starting near to a
good set of weights is important for fast convergence. An interesting
avenue for further exploration here is the effect of $\lambda$ on the
learning rate. Because the initial evaluation function is completely
wrong, there would be  some justification in setting $\lambda=1$ early
on so that KnightCap only tries to predict the outcome of the game and
not the evaluations of later moves (which are extremely unreliable). \\

\noindent {\bf Self-Play} \\ Learning by self-play was extremely
effective for TD-Gammon, but a significant reason for this is the
randomness of backgammon which ensures that with high probability
different games have substantially different sequences of moves, and
also the speed of play of TD-Gammon which ensured that learning could
take place over several hundred-thousand games.  Unfortunately, chess
programs are slow, and chess is a deterministic game, so self-play
by a deterministic algorithm tends to result in a large number of
substantially similar games. This is not a problem if the games seen
in self-play are ``representative'' of the games played in practice,
however KnightCap's self-play games with only non-zero material
weights are very different to the kind of games humans of the same
level would play.

To demonstrate that learning by self-play for KnightCap is not as
effective as learning against real opponents, we ran another
experiment in which all but the material parameters were initialised
to zero again, but this time KnightCap learnt by playing against
itself. After 600 games (twice as many as in the original FICS
experiment), we played the resulting version against the good version
that learnt on FICS for a further 100 games with the weight values
fixed. The self-play version scored only 11\% against the good FICS
version.

Simultaneously with the work presented here, Beal and Smith
\cite{beal97} reported positive results using essentially
TDLeaf($\lambda$) and self-play (with some random move choice) when
learning the parameters of an evaluation function that only computed
material balance.  However, they were not comparing performance
against on-line players, but were primarily investigating whether the
weights would converge to ``sensible'' values at least as good as the
naive (1, 3, 3, 5, 9) values for (pawn, knight, bishop, rook, queen)
(they did, within 2000 games, and using a value of $\lambda= 0.95$
which supports the discussion in ``good initial conditions'' above).

\section{Conclusion}
\label{conclusion}
We have introduced TDLeaf($\lambda$), a variant of TD($\lambda$)
suitable for training an evaluation function used in minimax
search. The only extra requirement of the algorithm is that the
leaf-nodes of the principal variations be stored throughout the game. 

We presented some experiments in which a chess evaluation function was
trained from B-grade to master level using TDLeaf($\lambda$) by
on-line play against a mixture of human and computer opponents. The
experiments show both the importance of ``on-line'' sampling (as
opposed to self-play) for a deterministic game such as chess, and the
need to start near a good solution for fast convergence, although just
how near is still not clear. 

On the theoretical side, it has recently been shown that TD($\lambda$)
converges for linear evaluation functions\cite{tsitsiklis97} (although
only in the sense of prediction, not control). An interesting avenue
for further investigation would be to determine whether
TDLeaf($\lambda$) has similar convergence properties.

\subsubsection*{Acknowledgements} 
Thanks to several of the anonymous referees for their helpful
remarks. Jonathan Baxter was supported by an Australian Postdoctoral
Fellowship. Lex Weaver was supported by an Australian Postgraduate
Research Award.

\bibliographystyle{abbrv}
\bibliography{bib}

\end{document}